\pdfoutput=1
\PassOptionsToPackage{x11names, svgnames}{xcolor}
\documentclass[11pt]{article}

\usepackage[final]{acl}

\usepackage{times}
\usepackage{latexsym}
\usepackage{tabularx}
\usepackage{booktabs}
\usepackage{array}
\usepackage{multirow}
\usepackage{graphicx}
\usepackage{amsmath}

\usepackage[breakable]{tcolorbox}

\usepackage[T1]{fontenc}

\usepackage[utf8]{inputenc}

\tcbset{
  remarkbox/.style={
    enhanced,
    colback=blue!5,
    colframe=blue!70!black,
    coltitle=white,
    fonttitle=\bfseries,
    title=Observation #1,
    boxrule=0.8pt,
    arc=3pt,
    outer arc=3pt,
    left=5pt,
    right=5pt,
    top=4pt,
    bottom=4pt
  }
}

\usepackage{listings}
\tcbuselibrary{skins,breakable}

\tcbset{
    promptbox/.style={
        colback=gray!5!white,
        colframe=gray!40!black,
        arc=2mm,
        boxrule=0.5pt,
        left=5mm,
        right=3mm,
        top=3mm,
        bottom=3mm,
        fonttitle=\bfseries,
        breakable
    }
}

\usepackage{microtype}

\usepackage{inconsolata}

\usepackage{graphicx}
\usepackage{enumitem}

\title{Thinking Out Loud: Do Reasoning Models Know When They're Right?}

\author{Qingcheng Zeng$^{1*}$, Weihao Xuan$^{2,3}$\thanks{Both authors contributed equally. Correspondence to \texttt{qcz@u.northwestern.edu}}, Leyang Cui$^{4}$, Rob Voigt$^{5}$\\\\
  $^{1}$Northwestern University, $^{2}$The University of Tokyo, $^{3}$RIKEN AIP\\$^{4}$Westlake University, $^{5}$University of California, Davis}

\begin{document}
\maketitle
\begin{abstract}
Large reasoning models (LRMs) have recently demonstrated impressive capabilities in complex reasoning tasks by leveraging increased test-time computation and exhibiting behaviors reminiscent of human-like self-reflection. While LRMs show a clear capacity for valuable self-reflection, how this ability interacts with other model behaviors remains underexplored. We investigate this connection by analyzing verbalized confidence, how models articulate their certainty, as a lens into the nature of self-reflection in LRMs. We find that supervised fine-tuning on reasoning traces (i.e., distillation) and reinforcement learning can improve verbalized calibration in reasoning-intensive settings in a progressive, laddered fashion. However, our results also indicate that reasoning models may possess a diminished awareness of their own knowledge boundaries, as evidenced by significantly lower ``\textit{I don't know}'' response rates on factuality benchmarks. Moreover, we examine the relationship between verbalized confidence and reasoning chains, finding that models tend to express higher confidence when providing shorter or less elaborate reasoning. Our findings highlight how reasoning-oriented training can enhance performance in reasoning-centric tasks while potentially incurring a \textit{reasoning tax}, a cost reflected in the model's reduced ability to accurately recognize the limits of its own knowledge in small-scale models. More broadly, our work showcases how this erosion of knowledge boundaries can compromise model faithfulness, as models grow more confident without a commensurate understanding of when they should abstain.
\end{abstract}

\section{Introduction}
\begin{figure}[!th]
    \centering
    \includegraphics[width=\columnwidth]{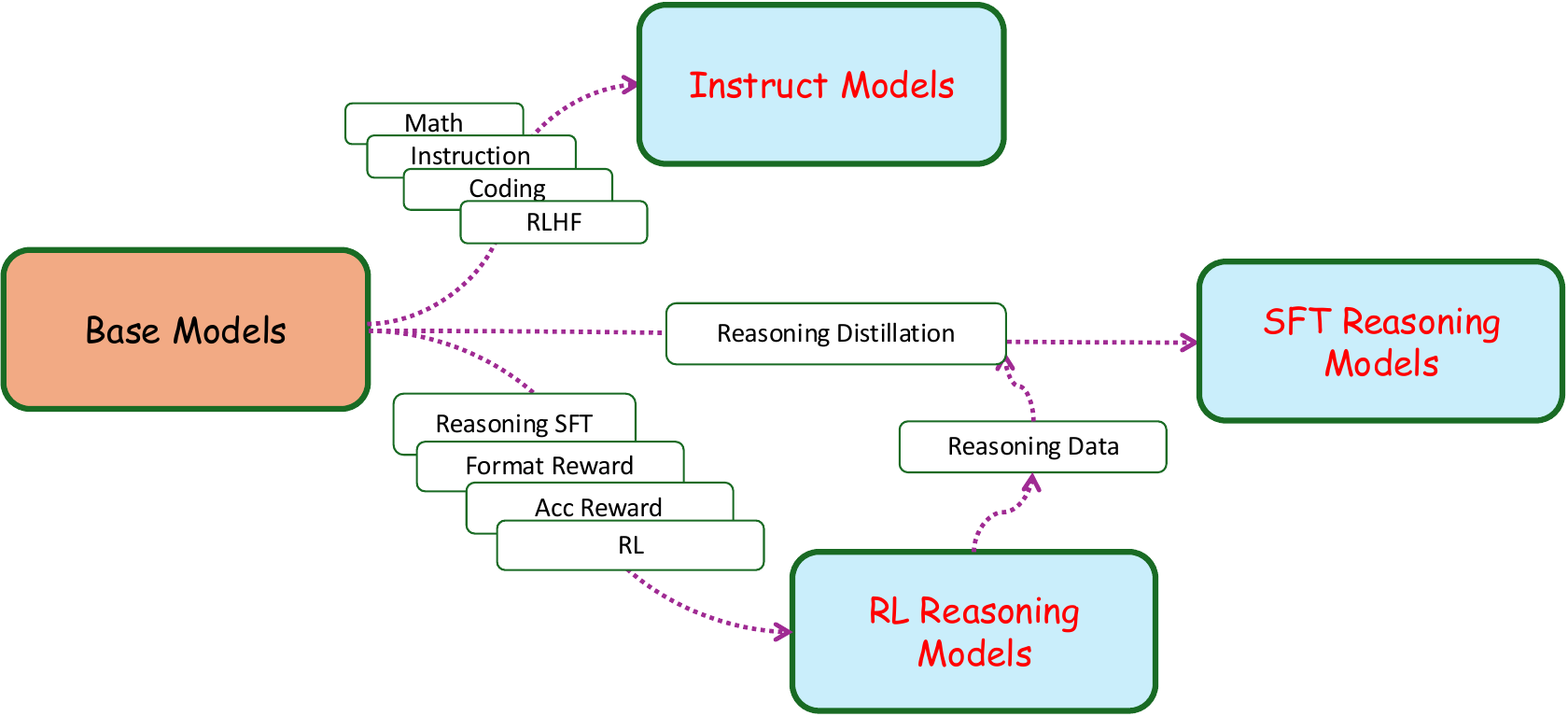}
    \caption{An illustration of different pathways of LLM/LRM training; we compare three key categories of models for their calibration performances.}
    \label{fig:illu}
\end{figure}

Large reasoning models (LRMs) have emerged as a dominant paradigm in the development of large language models (LLMs), achieving state-of-the-art performance across a range of complex tasks, including mathematics \cite{ye2025limoreasoning, moshkov2025aimo2}, complex reasoning \cite{openai2024openaio1card, deepseekai2025deepseekr1incentivizingreasoningcapability}, and coding \cite{openai2025competitiveprogramminglargereasoning}. A defining characteristic of LRMs is the emergence of self-reflective behaviors, where models appear capable of reassessing and refining their reasoning, sometimes displaying behavior suggestive of a nascent form of introspection that internally assesses whether particular stepwise inferences are likely correct or flawed (e.g. ``Wait, but...").

While verbalized uncertainty is widely used to assess calibration in LLMs \cite{wei2024measuringshortformfactualitylarge, phan2025humanitysexam, wei2025browsecompsimplechallengingbenchmark}, its association with the emerging self-reflective behaviors in LRMs remains underexplored. If reasoning models are truly more introspective, their verbalized confidence should better align with actual correctness. This motivates our central research question: Are reasoning models better calibrated? That is, do their improved reasoning capabilities lead to more faithful and reliable confidence estimates?

Prior studies have shown that LLMs often struggle to produce well-calibrated confidence estimates, frequently displaying overconfidence in their verbalized uncertainty. For example, \citet{xiong2024llmsexpressuncertaintyempirical} conducted a broad empirical study and found that many instruction-tuned LLMs systematically overstate their certainty across a variety of tasks, regardless of their actual correctness. Moreover, \citet{tian-etal-2023-just, xiong2024llmsexpressuncertaintyempirical, yang2024verbalizedconfidencescoresllms} have highlighted that model calibration is highly sensitive to prompt design, underscoring the fragility and lack of robustness in current approaches to verbalized uncertainty estimation in instruction-tuned models. While prior work suggests that human-inspired prompting strategies, such as chain-of-thought (CoT) or TopK, can enhance calibration, we extend this line of inquiry by investigating whether LRMs, which inherently embed long CoT chains and self-reflective behaviors, can further improve calibration.
In this work, we conduct a comprehensive empirical study to assess the calibration of LRMs across a diverse set of benchmarks spanning mathematics, factuality, scientific reasoning, and general reasoning. To isolate the effects of different training strategies, we evaluate models that share the same base architecture but vary in their post-training procedures. Our analysis focuses on three distinct model categories: (1) \textit{instruct models}, trained mainly using SFT and general RL for alignment purposes; (2) \textit{SFT reasoning models}, fine-tuned primarily on long CoT outputs generated by stronger reasoning models; and (3) \textit{RL reasoning models}, trained with reasoning RL to explicitly optimize reflective reasoning behaviors. An overview of these training pipelines is illustrated in Figure~\ref{fig:illu}. Through systematic pairwise comparisons, our key findings are as follows:
\begin{itemize}
    \item On reasoning-heavy benchmarks, both SFT reasoning models and RL reasoning models consistently outperform instruction-tuned models in terms of both task accuracy and calibration quality.
    \item While SFT on reasoning traces leads to substantial performance gains, RL offers additional improvements in calibration, even when the RL training domain (e.g., math) differs from the evaluation domain (e.g., science), highlighting its generalizability.
    \item On factuality-focused benchmarks, calibration improvements are less consistent: small-scale SFT reasoning models often exhibit worse calibration than instruction-tuned models, while RL reasoning models generally show some recovery. Further analysis indicates that open-source LRMs produce significantly fewer ``\textit{I don't know}'' responses compared to instruction-tuned models, suggesting a reduced awareness of their own knowledge boundaries.
\end{itemize}

\begin{figure*}[!th]
    \centering
    \includegraphics[width=0.85\textwidth]{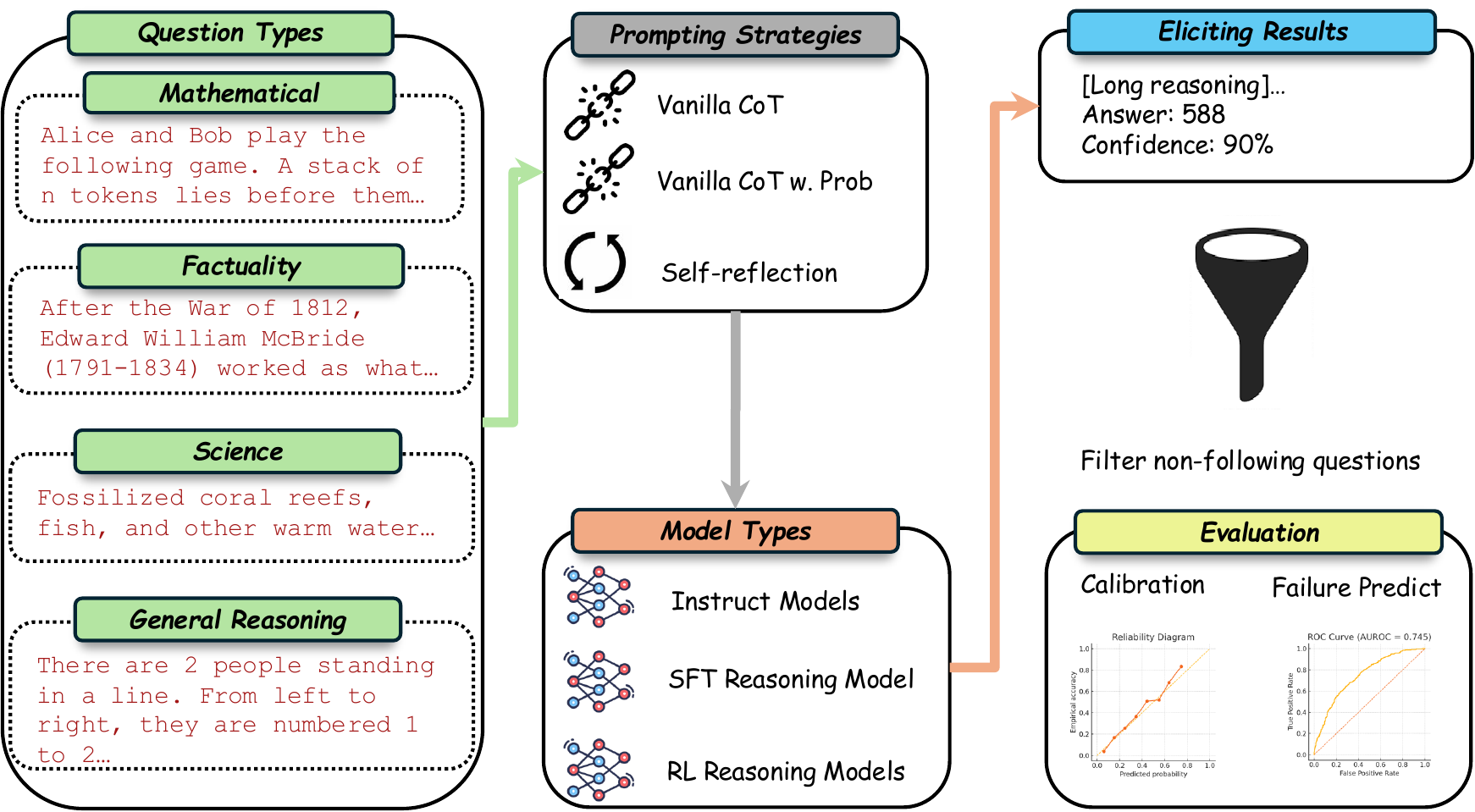}
    \caption{Verbalized confidence evaluation across various tasks, prompting strategies, and model types.}
    \label{fig:pipeline}
\end{figure*}

\section{Related Work}
\textbf{Large Reasoning Models.} Following the development of long-chain reasoning models \cite{ziabari2025reasoningspectrumaligningllms} such as o1 \cite{openai2024openaio1card}, a new generation of LLMs has emerged, designed to handle complex, multi-step reasoning tasks. The training of LRMs typically begins with an SFT phase on reasoning-intensive data, a process often referred to as a cold start. Even with limited curated data, this phase has shown strong results; for instance, \citet{ye2025limoreasoning} demonstrates that fine-tuning on just 817 human-curated examples can yield substantial improvements on mathematical and reasoning-heavy tasks. Building on this foundation, a second phase applies outcome-based RL to further enhance model performance by promoting self-exploration and reflective reasoning. Notable examples include DeepCoder-14B \cite{deepcoder2025}, QwQ \cite{qwq32b}, and DeepSeek-R1 \cite{deepseekai2025deepseekr1incentivizingreasoningcapability}, which use RL to refine the introspective and reasoning capabilities of LRMs beyond what SFT alone can achieve.

\textbf{Uncertainty Quantification.} Effectively quantifying uncertainty or confidence in LLMs is critical for assessing whether these models are inherently calibrated. \citet{vashurin2025benchmarkinguncertaintyquantificationmethods} provide a comprehensive benchmark of uncertainty quantification techniques across general-purpose LLMs, highlighting the strong performance of sampling-based methods, such as semantic entropy \cite{kuhn2023semanticuncertaintylinguisticinvariances} and SentenceSAR \cite{duan-etal-2024-shifting}, particularly in open-ended generation tasks. However, despite their accuracy, these approaches are computationally expensive, motivating increased interest in verbalized confidence as a lightweight alternative. For instance, \citet{tian-etal-2023-just} show that verbalized confidence can yield strong calibration in RLHF-trained models. \citet{xuan2025seeingbelievingmuchcomprehensive} explore how vision language models perform in multimodal settings and report that visual reasoning usually enhances better verbalized calibration. Yet, findings remain mixed: \citet{xiong2024llmsexpressuncertaintyempirical} report that many LLMs systematically overestimate their certainty, leading to poor calibration in practice. In this work, we revisit this issue by systematically comparing instruction-tuned models and reasoning models within a unified evaluation framework to examine how training paradigms affect verbalized calibration behavior.

\section{Experimental Setup}
\subsection{Models}
To evaluate the three model variants, we consider the following representative models within each category:

\begin{itemize}[itemsep=0pt, topsep=0pt]
    \item \textbf{Instruct models}: These include Qwen2.5-14B-Instruct, Qwen2.5-32B-Instruct (henceforth Qwen2.5-14/32B) \cite{qwen2025qwen25technicalreport}, and DeepSeek-V3 \cite{deepseekai2025deepseekv3technicalreport}. Specifically, these models were trained using SFT on both reasoning and non-reasoning data, then followed by general RL, mainly for alignment purposes.
    \item \textbf{SFT reasoning models}: We evaluate DeepSeek-R1-Distill-Qwen-14B/32B \cite{deepseekai2025deepseekr1incentivizingreasoningcapability}, both of which are fine-tuned on 800k examples (600k long reasoning chains and 200k non-reasoning data) distilled from the outputs of the intermediate DeepSeek-R1 model.
    \item \textbf{RL reasoning models}: This category includes DeepCoder-14B-Preview \cite{deepcoder2025}, Skywork-OR1-32B-Preview \cite{skywork-or1-2025}, and DeepSeek-R1 \cite{deepseekai2025deepseekr1incentivizingreasoningcapability}. We chose the former two models because they are trained on top of our evaluated SFT reasoning models, so that we could directly observe the reasoning RL's effects.
\end{itemize}

\subsection{Datasets}
We evaluate the models across the following datasets:

\begin{itemize}[itemsep=0pt, topsep=0pt] 
    \item \textbf{Math}: AIME 2024 and AIME 2025 \cite{aime}, each consisting of 30 challenging mathematical questions designed to test mathematical reasoning skills. We run both datasets five times and report the aggregated results.
    \item \textbf{Factuality}: SimpleQA \cite{wei2024measuringshortformfactualitylarge} and FreshQA-2025-04-28 \cite{vu2023freshllmsrefreshinglargelanguage}, two factuality benchmarks that evaluate the ability of language models to answer fact-seeking questions. In evaluation, we adopt the same prompt as in \citet{wei2024measuringshortformfactualitylarge} to categorize responses into correct, incorrect, or not attempted.
    \item \textbf{Scientific Reasoning}: GPQA-Diamond \cite{rein2023gpqagraduatelevelgoogleproofqa}, containing 198 graduate-level scientific multiple-choice questions; and SuperGPQA \cite{pteam2025supergpqascalingllmevaluation}, in which we randomly sample 500 questions from easy, medium, and hard levels, totaling 1500 questions.
    \item \textbf{General Reasoning}: The reasoning portion of LiveBench \cite{white2024livebenchchallengingcontaminationfreellm}, which we refer as LiveBench-Reasoning, contains 150 reasoning problems which are a harder version of Web of Lies from Big-Bench Hard \cite{suzgun2022challengingbigbenchtaskschainofthought} and Zebra Puzzles. We also run this dataset five times and report aggregated results.
\end{itemize}


Building on our experimental setup, we examine how reasoning-focused training affects model calibration across domains. Specifically, we test whether SFT on long reasoning traces improves calibration over general post-training and whether reasoning RL further enhances calibration. Finally, we assess whether these calibration gains transfer to less reasoning-focused domains like factuality, a key test of robustness and generalizability in real-world settings.

\subsection{Experimental Settings}
For model inference, we use the Huggingface Transformers library \cite{wolf2020huggingfacestransformersstateoftheartnatural} for all models except the DeepSeek variants, for which we rely on API-based inference. We consistently set the decoding temperature to \texttt{0.6} and allow up to 32,000 new tokens to ensure the generation of sufficiently detailed reasoning chains.

For prompting strategies, we evaluate the following approaches: (1) \textit{Vanilla chain-of-thought (CoT) prompting}: We use a slightly modified version of the method introduced by \citet{wei2024measuringshortformfactualitylarge}, incorporating a single CoT component to elicit reasoning from all models; (2) \textit{Vanilla CoT prompting with probability mass}: Motivated by \citet{yang2024verbalizedconfidencescoresllms}, who find that requesting confidence estimates as probability scores (ranging from 0.0 to 1.0) can improve calibration, we also test this approach; and (3) \textit{Self-reflection prompting}: This strategy uses a two-round dialogue, with the first round eliciting an answer and the second prompting the model to evaluate its own confidence. Detailed prompt templates are provided in \autoref{appendix:prompt}.

\subsection{Tasks and Metrics}
Leveraging the confidence scores elicited from LLMs, we investigate two complementary tasks: calibration and failure prediction \cite{Yuan_2021_ICCV, xiong2022birdsfeathertrusttogether}. Calibration assesses how well a model’s predicted confidence matches its actual accuracy. For example, a well-calibrated model should be correct 70\% of the time when it assigns 70\% confidence to its predictions. In contrast, failure prediction evaluates a model’s ability to distinguish between correct and incorrect predictions based on its confidence scores. Ideally, a model should assign higher confidence to correct answers and lower confidence to incorrect ones.

To quantify calibration performance, we use the Expected Calibration Error (ECE) \cite{pmlr-v70-guo17a}, which measures the average discrepancy between predicted confidence and empirical accuracy across bins. Specifically, ECE involves dividing samples into $M$ equal bins by confidence scores, then computing the mean absolute difference between each bin’s accuracy and average confidence: $\text{ECE} = \sum_{m=1}^{M} \frac{|B_m|}{n} \left| \text{acc}(B_m) - \text{avgConf}(B_m) \right|$, with $n$ as the total number of samples and $B_m$ as the set of samples in the $m$-th bin. Additionally, to address ECE’s sensitivity to binning strategies and its potential high variance, we also employ the Adaptive Calibration Error (ACE) \cite{nixon2019measuring}: $\text{ACE} = \frac{1}{M} \sum_{m=1}^{M} \left| \text{acc}(B_m) - \text{avgConf}(B_m) \right|$, which dynamically adjusts bin boundaries to ensure each bin contains an equal number of samples based on the data distribution. In all experiments, we use $M = 10$ bins for both ECE and ACE. Specifically, when we are evaluating calibration on factuality benchmarks, we only take attempted questions into calculation.

To assess how well confidence scores distinguish correct from incorrect predictions, we report the AUROC. We also include AUPRC for both positive and negative instances, as it offers complementary insight in imbalanced settings or when model accuracy varies. Finally, we report accuracy as a baseline measure of overall performance.

\begin{table*}[htbp]
\renewcommand{\arraystretch}{0.9}
\centering
\resizebox{\textwidth}{!}{%
\begin{tabular}{llcccccc}
\toprule
& & \textbf{Math} & \multicolumn{2}{c}{\textbf{Science Reasoning}} & \textbf{General Reasoning} & \multicolumn{2}{c}{\textbf{Factuality}} \\ 
\midrule
Metric & Model & AIME 2024 \& 2025 & GPQA-Diamond & SuperGPQA & LiveBench-Reasoning & SimpleQA & FreshQA \\
\midrule
\multirow{7}{*}{Acc $\uparrow$} 
    & \textcolor{Orange}{Qwen2.5-14B}             & 11.3\%   & 35.8\%   & 29.3\%  & 38\% & 6.04\% & 38.3\% \\
    & \textcolor{RoyalBlue}{R1-Distill-Qwen-14B} & 46.7\% & 54.0\% & 40.67\% & 58.7\% & 5.69\% & 32.2\% \\
    & \textcolor{Crimson}{DeepCoder-14B} & 57.7\% & 56.1\% & 41.4\% & 62.7\% & 5.28\% & 32.7\% \\
    & \textcolor{Orange}{Qwen2.5-32B} & 9.67\% & 39.4\% & 31.2\% & 42.7\% & 5.32\% & 35.2\% \\
    & \textcolor{RoyalBlue}{R1-Distill-Qwen-32B} & 65.7\% & 62.6\% & 48.1\% & 73.3\% & 7.28\% & 36.3\% \\
    & \textcolor{Crimson}{Skywork-32B} & 51.3\% & 63.6\% & 50.4\% & 84.7\% & 6.80\% & 36.2\% \\
    & \textcolor{Orange}{DeepSeek-V3} & 23.0\% & 48.5\% & 39.3\% & 50.0\% & 21.4\% & 52.4\% \\
    & \textcolor{Crimson}{DeepSeek-R1} & 68.0\% & 68.7\% & 60.3\% & 89.3\% & 29.7\% & 53.5\% \\
\midrule
\multirow{7}{*}{ECE/ACE $\downarrow$}
    & \textcolor{Orange}{Qwen2.5-14B}  & 0.760/0.759  & 0.469/0.466    & 0.514/0.511  & 0.540/0.536 & 0.625/0.625 & 0.436/0.432 \\
    & \textcolor{RoyalBlue}{R1-Distill-Qwen-14B} & 0.342/0.342 & 0.244/0.243 & 0.386/0.385 & 0.265/0.285 & 0.719/0.719 & 0.523/0.525 \\
    & \textcolor{Crimson}{DeepCoder-14B} & 0.222/0.227 & 0.225/0.233 & 0.378/0.377 & 0.222/0.227 & 0.705/0.705 & 0.514/0.514 \\
    & \textcolor{Orange}{Qwen2.5-32B} & 0.752/0.751 & 0.411/0.406 & 0.446/0.444 & 0.472/0.472 & 0.623/0.622 & 0.438/0.440 \\
    & \textcolor{RoyalBlue}{R1-Distill-Qwen-32B} & 0.240/0.240 & 0.217/0.234 & 0.352/0.352 & 0.152/0.162 & 0.702/0.702 & 0.483/0.485 \\
    & \textcolor{Crimson}{Skywork-32B} & 0.183/0.188 & 0.174/0.179 & 0.298/0.293 & 0.074/0.053 & 0.624/0.623 & 0.442/0.446 \\
    & \textcolor{Orange}{DeepSeek-V3} & 0.570/0.572 & 0.354/0.357 & 0.427/0.424 & 0.389/0.389 & 0.515/0.515 & 0.356/0.358 \\
    & \textcolor{Crimson}{DeepSeek-R1} & 0.136/0.142 & 0.082/0.094 & 0.160/0.156 & 0.081/0.081 & 0.324/0.324 & 0.299/0.300 \\
\midrule
\multirow{7}{*}{AUROC $\uparrow$}
    & \textcolor{Orange}{Qwen2.5-14B}  & 0.670  & 0.637    & 0.597  & 0.489 & 0.622 & 0.726 \\
    & \textcolor{RoyalBlue}{R1-Distill-Qwen-14B} & 0.847 & 0.737 & 0.633 & 0.766 & 0.613 & 0.754 \\
    & \textcolor{Crimson}{DeepCoder-14B} & 0.873 & 0.779 & 0.627 & 0.797 & 0.632 & 0.738 \\
    & \textcolor{Orange}{Qwen2.5-32B} & 0.695 & 0.603 & 0.644 & 0.556 & 0.615 & 0.732 \\
    & \textcolor{RoyalBlue}{R1-Distill-Qwen-32B} & 0.813 & 0.798 & 0.659 & 0.777 & 0.611 & 0.769 \\
    & \textcolor{Crimson}{Skywork-32B} & 0.928 & 0.790 & 0.665 & 0.876 & 0.615 & 0.789 \\
    & \textcolor{Orange}{DeepSeek-V3} & 0.798 & 0.719 & 0.645 & 0.696 & 0.695 & 0.740 \\
    & \textcolor{Crimson}{DeepSeek-R1} & 0.942 & 0.793 & 0.657 & 0.908 & 0.705 & 0.767 \\
\midrule
\multirow{7}{*}{AUPRC-P/AUPRC-N $\uparrow$}
    & \textcolor{Orange}{Qwen2.5-14B} & 0.170/0.941  & 0.449/0.731 & 0.423/0.768 & 0.381/0.600 & 0.110/0.949 & 0.629/0.765 \\
    & \textcolor{RoyalBlue}{R1-Distill-Qwen-14B} & 0.816/0.856 & 0.742/0.659 & 0.536/0.685 & 0.794/0.654 & 0.096/0.958 & 0.615/0.847 \\
    & \textcolor{Crimson}{DeepCoder-14B} & 0.895/0.846 & 0.821/0.672 & 0.535/0.678 & 0.815/0.690 & 0.094/0.965 & 0.552/0.822 \\
    & \textcolor{Orange}{Qwen2.5-32B} & 0.135/0.959 & 0.489/0.668 & 0.442/0.776 & 0.467/0.632 & 0.159/0.917 & 0.628/0.785 \\
    & \textcolor{RoyalBlue}{R1-Distill-Qwen-32B} & 0.870/0.753 & 0.868/0.630 & 0.641/0.642 & 0.896/0.476 & 0.111/0.949 & 0.676/0.821 \\
    & \textcolor{Crimson}{Skywork-32B} & 0.957/0.870 & 0.870/0.639 & 0.656/0.625 & 0.974/0.387 & 0.100/0.954 & 0.658/0.858 \\
    & \textcolor{Orange}{DeepSeek-V3} & 0.442/0.933 & 0.648/0.711 & 0.520/0.734 & 0.663/0.690 & 0.363/0.879 & 0.737/0.724 \\
    & \textcolor{Crimson}{DeepSeek-R1} & 0.960/0.906 & 0.896/0.543 & 0.714/0.538 & 0.974/0.498 & 0.499/0.843 & 0.765/0.748 \\
\bottomrule
\end{tabular}
}%
\caption{Performance metrics for all models using the vanilla CoT prompting strategy. Accuracy (Acc) reflects task performance; ECE/ACE, AUROC, and AUPRC-P/N assess calibration and failure prediction. Fewer than 1.5\% of instances did not follow instructions and are excluded from analysis. Colors indicate model types: orange for instruct, blue for SFT, and red for RL reasoning models.}
\label{tab:results_metrics}
\end{table*}

\section{Results}
The results of our main evaluation are presented in Table~\ref{tab:results_metrics}. As a sanity check, we observe that the overall performance closely aligns with the results reported in prior work \cite{deepseekai2025deepseekr1incentivizingreasoningcapability, deepcoder2025}. This consistency indicates that the inclusion of confidence elicitation alongside answer generation does not substantially affect the models' general performance.

\subsection{General Results}
\begin{tcolorbox}[remarkbox=1]
SFT on reasoning data significantly improves both accuracy and calibration in reasoning-dense scenarios.
\end{tcolorbox}

\begin{figure}[!th]
    \centering
    \includegraphics[width=\columnwidth]{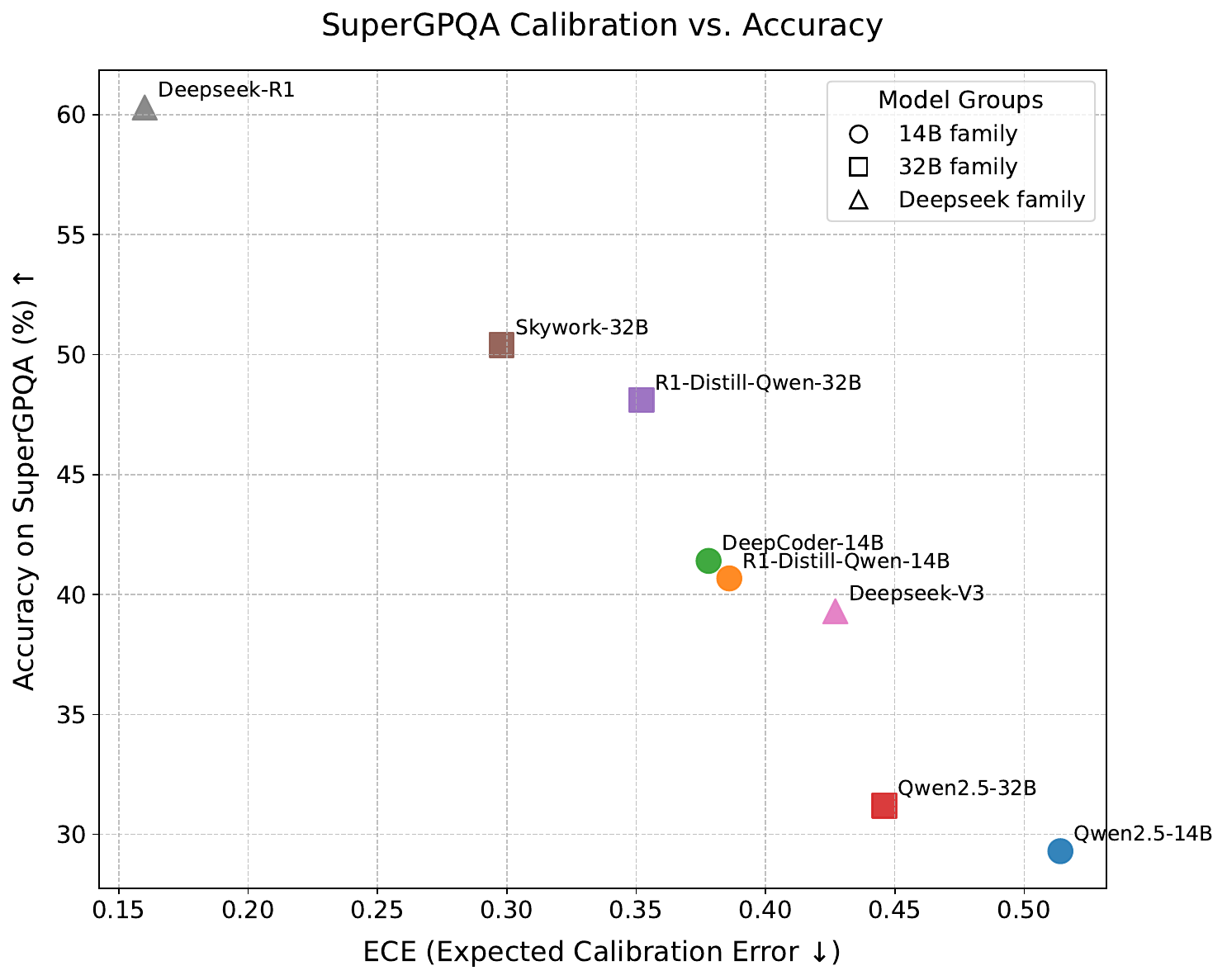}
    \caption{The relationship between accuracy and ECE on SuperGPQA benchmark. Upper left represents better models.}
    \label{fig:supergpqa}
\end{figure}

To evaluate the effect of SFT on reasoning data, we compare SFT and reasoning SFT variants of Qwen2.5 at both 14B and 32B model scales. The results reveal a consistent and notable trend: fine-tuning on long-form reasoning traces substantially improves task accuracy, while also leading to markedly better calibration, reflected in lower ECE and ACE scores. For example, R1-Distill-Qwen-32B improves accuracy on AIME from 9.67\% to 65.7\%, reduces ECE from 0.752 to 0.240, and boosts AUROC from 0.695 to 0.813, indicating that the model not only becomes more capable but also more aligned in its confidence assessments.

However, we also observe an intriguing side effect: AUPRC-N (which quantifies how well the model can distinguish incorrect answers) often declines following SFT. This suggests that while the model becomes more accurate and confident, its errors become less separable by confidence level, possibly because incorrect predictions now occur in harder or more ambiguous cases, where the model remains relatively confident. This highlights a trade-off where SFT enhances overall capability and confidence calibration, but may obscure signals useful for failure prediction.

Despite this nuance, the overall benefits are clear. Reasoning-oriented SFT significantly improves both task performance and calibration in reasoning-heavy scenarios, and these gains are consistent across different model scales. This indicates that SFT on reasoning traces provides a scalable and effective approach for improving not just accuracy, but also the reliability of verbalized uncertainty in LRMs.

\begin{tcolorbox}[remarkbox=2]
Reasoning RL provides additional calibration and performance benefits beyond SFT, even in the presence of domain mismatch.
\end{tcolorbox}

Beyond the improvements achieved through reasoning-oriented SFT, RL further enhances both model performance and calibration. When comparing RL reasoning models such as DeepCoder-14B and Skywork-32B to their SFT-only counterparts, we observe consistent gains in calibration and failure prediction metrics, including lower ECE and higher AUROC scores. While accuracy gains are somewhat task-dependent, the calibration advantage of RL reasoning models is robust across all evaluated reasoning benchmarks. Notably, although DeepSeek-R1 is not a direct RL continuation of DeepSeek-V3, the contrast between the two provides additional evidence that RL-style training meaningfully improves the alignment between confidence and correctness. As shown in \autoref{fig:supergpqa}, we witness a clear trend of the gains from instruct to SFT and finally RL models.

These findings suggest that RL serves as a valuable complement to SFT, encouraging models to develop not only stronger reasoning capabilities but also more trustworthy self-assessment. Furthermore, the fact that DeepCoder and Skywork were fine-tuned with RL on different domains (coding and mathematics, respectively) and developed by independent organizations, yet still exhibit calibration improvements across a wide range of tasks, supports the view that RL-enhanced calibration generalizes across domains. More broadly, this highlights RL as a promising direction for aligning LLMs' verbalized confidence with actual reliability, a key requirement for deploying these systems in high-stakes or decision-making applications.

\subsection{A Deep Look into the Factuality Benchmark}
\begin{tcolorbox}[remarkbox=3]
Reasoning models usually show significantly lower “\textit{not attempted}” responses with non-significant accuracy improvement.
\end{tcolorbox}

While reasoning-oriented training enhances both performance and calibration on complex reasoning tasks, our analysis reveals a potential drawback in domains that demand factual precision and less reasoning. As shown in the performances of SimpleQA and FreshQA, small-scale reasoning models generally exhibit lower calibration compared to instruction-tuned models, though RL reasoning models show a slight improvement over SFT counterparts. 

To further investigate this, we first report the number of “\textit{not attempted}” responses across the models we evaluate, as shown in \autoref{tab:not_attempted}. Our results indicate that LRMs usually exhibit significantly lower rates of “\textit{I don't know}” responses compared to instruction-tuned models, which were trained with general-purpose RL for alignment \footnote{Given our evaluated reasoning models start from the base model, we also test SFT reasoning models which start from instruct models. Details are attached in \autoref{appendix:further_factuality}.}. However, despite this reduced hesitation, LRMs do not consistently achieve a significantly higher accuracy on factuality benchmarks except DeepSeek-R1. Taken together, these results suggest that small-scale LRMs might have a diminished ability to recognize the limits of their own knowledge.

\begin{table}[htbp]
\centering
\resizebox{\columnwidth}{!}{%
\begin{tabular}{lccc}
\toprule
Model Size & Instruct & SFT Reasoning & RL Reasoning \\
\midrule
14B & 1136 & 102 & 103 \\
32B & 2492 & 76 & 63 \\
DeepSeek & 480 & - & 81 \\
\bottomrule
\end{tabular}
}
\caption{The total number of “\textit{not attempted}” responses in SimpleQA and FreshQA.}
\label{tab:not_attempted}
\end{table}


\begin{table*}[t]
\small
\centering
\renewcommand{\arraystretch}{0.75}
\resizebox{0.7\textwidth}{!}{%
{\small
\begin{tabular}{lllccc}
\toprule
\multirow{2}{*}{Question Categories} & \multirow{2}{*}{Model Size} & \multirow{2}{*}{Metric} & \multirow{2}{*}{Instruct} & \multirow{2}{*}{SFT Reasoning} & \multirow{2}{*}{RL Reasoning} \\
                          &                        &                         &          &               &              \\
\midrule
\multirow{6}{*}{Shared Attempted} 
  & \multirow{2}{*}{14B}      & Acc &    12.5\%    &    10.5\%    &    9.98\%    \\
  &                           & ECE &    0.598     &     0.692    &     0.684   \\
  & \multirow{2}{*}{32B}      & Acc &    17.4\%    &    17.5\%    &     17.1\%   \\
  &                           & ECE &    0.591    &     0.640  &      0.600  \\
  & \multirow{2}{*}{DeepSeek} & Acc &   27.5\%     &    -    &  34.6\%      \\
  &                           & ECE &    0.496    &     -   &  0.317      \\
\midrule
\multirow{6}{*}{Only LRMs Attempted} 
  & \multirow{2}{*}{14B}      & Acc &    0\%    &    2.37\%    &    2.75\%    \\
  &                           & ECE &    -    &    0.718    &    0.690    \\
  & \multirow{2}{*}{32B}      & Acc &    0\%    &    3.7\%    &    3.23\%    \\
  &                           & ECE &    -    &    0.717    &   0.653     \\
  & \multirow{2}{*}{DeepSeek} & Acc &    0\%    &    -    &    11.4\%    \\
  &                           & ECE &    -    &    -    &  0.371      \\
\bottomrule
\end{tabular}%
}}
\caption{Factuality evaluation results for two question categories across instruct, SFT reasoning, and RL reasoning models. Instruct models do not have ECE for not attempted responses.}
\label{tab:factuality_deep}
\end{table*}

In \autoref{tab:factuality_deep}, we delve deeper into our factuality benchmarks by analyzing two types of questions: (1) questions that are answered by both instruction-tuned and reasoning models of the same scale (shared questions), and (2) questions that are not attempted by instruction-tuned models but are answered by same-scale reasoning models. Our results show that, for shared questions, smaller reasoning models generally do not achieve notable accuracy gains and often exhibit worse calibration, particularly in the case of SFT reasoning models. A similar trend is observed in the second category, where smaller reasoning models attempt additional questions but achieve only marginal accuracy and display relatively high calibration error.

In contrast, larger reasoning models such as DeepSeek demonstrate clear performance gains in both categories. Notably, they also show improved calibration on shared questions, indicating that larger-scale models benefit more from reasoning-focused training, both in terms of capability and confidence alignment. These findings suggest that reasoning RL plays an important role in producing more reliable verbalized uncertainty, even in factuality-focused tasks where reasoning is less central.

\subsection{Do Prompting Strategies Matter?}
\begin{table*}[htbp]
\centering
\resizebox{\textwidth}{!}{%
\begin{tabular}{lc|ccc|ccc|ccc|ccc|ccc|ccc}
\toprule
Metric & Model 
& \multicolumn{3}{c|}{AIME 2024 \& 2025} 
& \multicolumn{3}{c|}{GPQA-Diamond} 
& \multicolumn{3}{c|}{SuperGPQA} 
& \multicolumn{3}{c|}{LiveBench-Reasoning} 
& \multicolumn{3}{c|}{SimpleQA} 
& \multicolumn{3}{c}{FreshQA} \\
\cmidrule(lr){3-5} \cmidrule(lr){6-8} \cmidrule(lr){9-11} \cmidrule(lr){12-14} \cmidrule(lr){15-17} \cmidrule(lr){18-20}
& 
& V & V.Prob & SR 
& V & V.Prob & SR 
& V & V.Prob & SR
& V & V.Prob & SR 
& V & V.Prob & SR 
& V & V.Prob & SR \\
\midrule
\multirow{8}{*}{ECE $\downarrow$}
& Qwen2.5-14B             & 0.760 & 0.768 & 0.422 & 0.469 & 0.508 & 0.479 & 0.514 & 0.531 & 0.474 & 0.540 & 0.481 & 0.443 & 0.625 & 0.664 & 0.515 & 0.436 & 0.432 & 0.287 \\
& R1-Distill-Qwen-14B     & 0.342 & 0.305 & 0.223 & 0.244 & 0.235 & 0.244 & 0.386 & 0.421 & 0.420 & 0.265 & 0.228 & 0.199 & 0.719 & 0.709 & 0.738 & 0.523 & 0.530 & 0.477 \\
& DeepCoder-14B           & 0.222 & 0.260 & 0.244 & 0.225 & 0.220 & 0.255 & 0.378 & 0.400 & 0.425 & 0.222 & 0.225 & 0.227 & 0.705 & 0.696 & 0.734 & 0.514 & 0.521 & 0.465 \\
& Qwen2.5-32B             & 0.752 & 0.740 & 0.289 & 0.411 & 0.385 & 0.382 & 0.446 & 0.422 & 0.382 & 0.472 & 0.496 & 0.473 & 0.623 & 0.614 & 0.506 & 0.438 & 0.396 & 0.389 \\
& R1-Distill-Qwen-32B     & 0.240 & 0.223 & 0.193 & 0.217 & 0.257 & 0.189 & 0.352 & 0.374 & 0.370 & 0.152 & 0.213 & 0.100 & 0.702 & 0.700 & 0.725 & 0.483 & 0.492 & 0.439 \\
& Skywork-32B             & 0.183 & 0.195 & 0.076 & 0.174 & 0.192 & 0.208 & 0.298 & 0.292 & 0.338 & 0.074 & 0.037 & 0.080 & 0.624 & 0.617 & 0.656 & 0.442 & 0.442 & 0.370 \\
& DeepSeek-V3             & 0.570 & 0.562 & 0.502 & 0.354 & 0.368 & 0.379 & 0.427 & 0.414 & 0.413 & 0.389 & 0.308 & 0.309 & 0.515 & 0.505 & 0.500 & 0.356 & 0.376 & 0.338 \\
& DeepSeek-R1             & 0.136 & 0.142 & 0.167 & 0.082 & 0.074 & 0.119  &  0.160   & 0.162 & 0.265 &  0.081  & 0.071 & 0.077 & 0.324 & 0.305 & 0.551 & 0.301 & 0.294 & 0.324 \\
\bottomrule
\end{tabular}
}
\caption{ECE ($\downarrow$) of models across datasets and prompting strategies. Here, V stands for vanilla CoT, V.Prob stands for vanilla CoT with probability mass, and SR stands for self-reflection.}
\label{tab:ece_prompting_strategies}
\end{table*}

We present the ECE results of the three prompting strategies in \autoref{tab:ece_prompting_strategies}. On reasoning benchmarks, RL reasoning models, regardless of model size, consistently achieve the best calibration across all prompting strategies. This finding highlights the stable and robust effect of reasoning RL in improving the alignment between model confidence and accuracy. In contrast, on factuality benchmarks, smaller-scale reasoning models tend to be more miscalibrated than instruction-tuned models, a pattern that persists across different prompting strategies. Notably, it is only among large-scale reasoning models, such as DeepSeek, that we observe consistently improved calibration. This pattern reinforces the idea that both model scale and the application of RL training paradigms play a critical role in achieving generalizable, well-calibrated confidence estimates.

Interestingly, we observe divergent effects of self-reflection (SR) prompting in factuality-focused tasks. In SimpleQA, SR often harms calibration, increasing model overconfidence. Conversely, in FreshQA, SR generally improves calibration, particularly for smaller models. This contrast suggests that the utility of SR prompting may be influenced by dataset-specific characteristics, such as the prevalence of false premises in FreshQA or overall task difficulty. Taken together, these findings indicate that while prompting strategies like SR can modulate calibration in certain contexts, the dominant factors shaping verbalized uncertainty remain the model’s training paradigm and scale, especially the inclusion of RL-based objectives.

\subsection{A Deep Look Into the Length of Reasoning Chains}
\begin{figure*}[!th]
    \centering
    \includegraphics[width=\textwidth]{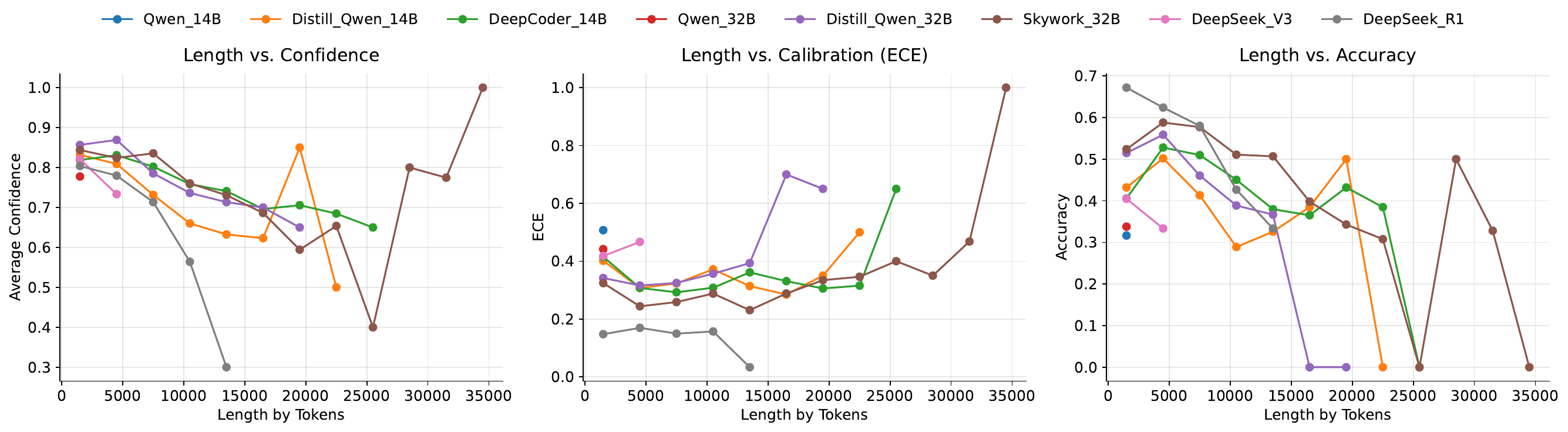}
    \caption{The relationship between length and confidence, calibration, and accuracy on GPQA-Diamond and SuperGPQA benchmarks.}
    \label{fig:thoughtology}
\end{figure*}

In this section, inspired by the concept of Thoughtology \cite{marjanović2025deepseekr1thoughtologyletsthink}, we analyze the relationship between reasoning chain length and model behavior, focusing on accuracy, verbalized confidence, and calibration (measured by ECE). These results are visualized in \autoref{fig:thoughtology}, using our Science QA benchmarks as the testbed.

Consistent with the findings of \citet{marjanović2025deepseekr1thoughtologyletsthink}, we observe that longer reasoning chains are generally associated with lower accuracy. In our analysis, this decline in accuracy is also accompanied by a reduction in verbalized confidence, suggesting that models may internally register when they are failing to answer well as they generate extended reasoning traces. This effect is especially pronounced in DeepSeek-R1, where we observed a significant drop in confidence for longer chains.

However, the relationship between reasoning chain length and calibration is less straightforward. For reasoning chains shorter than 10,000 tokens, ECE remains relatively stable, with no clear trend of improvement or degradation. When chain length exceeds this threshold, we observe a modest increase in ECE for small-scale models, implying that extremely long reasoning chains may introduce additional uncertainty or overconfidence that models do not appropriately adjust for. In DeepSeek-R1, we did not observe excessively long reasoning chains, and ECE remained lower even when surpassing 10,000 tokens, likely due to the more pronounced drop in model confidence.

\section{Discussion}
Our results show that reasoning-oriented training strategies improve performance on complex reasoning tasks. However, their effects on confidence calibration, particularly in factuality-focused benchmarks, are inconsistent. Concurrent works \cite{mei2025reasoninguncertaintyreasoningmodels, yoon2025reasoningmodelsbetterexpress} also report similar results on different sets of models.

\citet{wei2024measuringshortformfactualitylarge} compared the calibration of OpenAI’s o1 model with GPT-4o \cite{openai2024gpt4ocard} on factuality tasks, finding that o1 exhibited both better calibration and a higher rate of “\textit{not attempted}” responses. These results both align with and diverge from our observations. On one hand, we find that open-source LRMs attempt a substantially higher proportion of questions than instruction-tuned models, suggesting a reduced ability to recognize the limits of their knowledge, a potential weakness in current open-source LRM training pipelines. On the other hand, our results confirm that RL reasoning models, such as DeepSeek-R1, show improved calibration relative to their instruction-tuned counterparts, consistent with the o1-vs-GPT-4o comparison. However, reasoning-based SFT alone often leads to degraded calibration on factual benchmarks when compared to instruction-tuned baselines. Interestingly, we observe a partial recovery in calibration performance on factual benchmarks for RL reasoning models, revealing a ``U-shaped” trajectory in calibration quality across training paradigms: from instruction tuning, to reasoning SFT, to RL.

These findings contribute to ongoing discussions of the ``hallucination tax'' in reinforcement fine-tuning for reasoning performance. Concurrent works \cite{song2025hallucinationtaxreinforcementfinetuning, kirichenko2025abstentionbenchreasoningllmsfail} report that, after reinforcement learning, LRMs attempt answers substantially more often, even on intrinsically unanswerable questions. The authors attribute this ``hallucination tax'' in LRMs to reward misspecification when abstention would be appropriate. Likewise, \citet{kalai2025languagemodelshallucinate} argues that prevailing training and evaluation protocols reward guessing over acknowledging uncertainty, thereby amplifying hallucination. Taken together, these results underscore the need for more comprehensive evaluation protocols and reward designs for reinforcement learning systems \cite{tu2025positionhiddencostsmeasurement}.

Besides, our findings also add to ongoing discussions about the distinct roles of SFT and RL in shaping model generalization. \citet{chu2025sftmemorizesrlgeneralizes} describe SFT as a process that “\textit{memorizes},” while RL “\textit{generalizes}.” Our results refine this distinction: reasoning-based SFT improves in-domain calibration for complex reasoning tasks but may undermine calibration in domains requiring factual precision. In contrast, RL appears to support the development of more reflective and domain-agnostic confidence estimation, helping models slightly recover their verbalized uncertainty with correctness, even outside the primary distribution of their training data. These insights underscore the importance of balancing capability improvements with faithful self-assessment, especially as LLMs are deployed in increasingly open-ended and high-stakes environments.

Our results suggest that RL improves the calibration of verbalized uncertainty. Unlike sampling-based or post-hoc methods, verbalized confidence provides a natural, interpretable interface for human-AI interaction, allowing users to assess model certainty directly. In this context, calibration becomes essential for trustworthy deployment. We find that RL-trained models show more consistent alignment between expressed confidence and actual correctness, likely due to RL’s ability to foster reflective behavior beyond what SFT offers. As LLMs are increasingly deployed in high-stakes settings, reliable verbalized uncertainty is crucial for effective human-model collaboration.

\section*{Limitations}
This paper has two main limitations. First, despite prior work showing that reasoning models can handle code reasoning effectively \cite{openai2025competitiveprogramminglargereasoning}, we found that most models struggled to output both code snippets and confidence simultaneously. As a result, we excluded code reasoning from our evaluation. Second, while we follow the same evaluation procedure as \citet{wei2024measuringshortformfactualitylarge} for factuality benchmarks, we do not include human verification of the outputs.

\bibliography{custom}

\appendix

\section{Prompt Template}
\label{appendix:prompt}
\begin{tcolorbox}[promptbox, title=vanilla\_aime\_prompt\_template]
\small
\textbf{Task:} Solve the following math problem. Provide your best guess along with a confidence score (0\% to 100\%).
\\\\
\textbf{Instructions:}\\
- Please reason step by step.\\
- At the end, present your final answer and a confidence score in the following XML format:\\
<answer>final answer here</answer>\\
<confidence>confidence score here</confidence>\\
\\
\textbf{Example output:} \\
\char91 YOUR\_REASONING\char93\\
<answer>123</answer>\\
<confidence>80\%</confidence>\\
\\
\textbf{Now, here is the problem:} \\
\{problem\}
\end{tcolorbox}

\begin{tcolorbox}[promptbox, title=vanilla\_mc\_prompt\_template]
\small
\textbf{Task:} Solve the following multiple-choice problem. Provide your best guess along with a confidence score (0\% to 100\%).
\\\\
\textbf{Instructions:}\\
- Carefully read and analyze the problem.\\
- Reason through the solution step by step, if helpful.\\
- At the end, present your final answer and a confidence score in the following XML format:\\
<answer>final answer here</answer>\\
<confidence>confidence score here</confidence>\\
\\
\textbf{Example output:} \\
\char91 YOUR\_REASONING\char93\\
<answer>A</answer>\\
<confidence>80\%</confidence>\\
\\
\textbf{Now, here is the problem:} \\
\{problem\}
\end{tcolorbox}

\begin{tcolorbox}[promptbox, title=vanilla\_simpleqa\_prompt\_template]
\small
\textbf{Task:} Solve the following QA problem. Provide your best guess along with a confidence score (0\% to 100\%).
\\\\
\textbf{Instructions:}\\
- Carefully read and analyze the problem.\\
- Reason through the solution step by step, if helpful.\\
- At the end, present your final answer and a confidence score in the following XML format:\\
<answer>final answer here</answer>\\
<confidence>confidence score here</confidence>\\
\\
\textbf{Example output:} \\
\char91 YOUR\_REASONING\char93\\
<answer>123</answer>\\
<confidence>80\%</confidence>\\
\\
\textbf{Now, here is the problem:} \\
\{problem\}
\end{tcolorbox}

\begin{tcolorbox}[promptbox, title=livebench\_reasoning\_prompt\_template]
\small
\textbf{Task:} Solve the following reasoning problem. Provide your best guess along with a confidence score (0\% to 100\%).
\\\\
\textbf{Instructions:}\\
- Carefully read and analyze the problem.\\
- Reason through the solution step by step, if helpful.\\
- You might see several questions in the problem. You need to answer all of them and provide your final answer separated by commas.\\
- At the end, present your final answer and a confidence score in the following XML format:\\
<answer>final answer here</answer>\\
<confidence>confidence score here</confidence>\\
\\
\textbf{Example output:} \\
\char91 YOUR\_REASONING\char93\\
<answer>no, yes, no</answer>\\
<confidence>80\%</confidence>\\
\\
\textbf{Now, here is the problem:} \\
\{problem\}
\end{tcolorbox}

\begin{tcolorbox}[promptbox, title=vanilla\_aime\_prob\_prompt\_template]
\small
\textbf{Task:} Solve the following math problem. Provide your best guess along with a confidence probability score (0.0 to 1.0).
\\\\
\textbf{Instructions:}\\
- Please reason step by step.\\
- At the end, present your final answer and a confidence probability score in the following XML format:\\
<answer>final answer here</answer>\\
<confidence>confidence probability score here</confidence>\\
\\
\textbf{Example output:} \\
\char91 YOUR\_REASONING\char93\\
<answer>123</answer>\\
<confidence>0.8</confidence>\\
\\
\textbf{Now, here is the problem:} \\
\{problem\}
\end{tcolorbox}

\begin{tcolorbox}[promptbox, title=vanilla\_mc\_prob\_prompt\_template]
\small
\textbf{Task:} Solve the following multiple-choice problem. Provide your best guess along with a confidence probability score (0.0 to 1.0).
\\\\
\textbf{Instructions:}\\
- Carefully read and analyze the problem.\\
- Reason through the solution step by step, if helpful.\\
- At the end, present your final answer and a confidence probability score in the following XML format:\\
<answer>final answer here</answer>\\
<confidence>confidence probability score here</confidence>\\
\\
\textbf{Example output:} \\
\char91 YOUR\_REASONING\char93\\
<answer>A</answer>\\
<confidence>0.8</confidence>\\
\\
\textbf{Now, here is the problem:} \\
\{problem\}
\end{tcolorbox}

\begin{tcolorbox}[promptbox, title=vanilla\_simpleqa\_prob\_prompt\_template]
\small
\textbf{Task:} Solve the following QA problem. Provide your best guess along with a confidence probability score (0.0 to 1.0).
\\\\
\textbf{Instructions:}\\
- Carefully read and analyze the problem.\\
- Reason through the solution step by step, if helpful.\\
- At the end, present your final answer and a confidence probability score in the following XML format:\\
<answer>final answer here</answer>\\
<confidence>confidence probability score here</confidence>\\
\\
\textbf{Example output:} \\
\char91 YOUR\_REASONING\char93\\
<answer>123</answer>\\
<confidence>0.8</confidence>\\
\\
\textbf{Now, here is the problem:} \\
\{problem\}
\end{tcolorbox}

\begin{tcolorbox}[promptbox, title=livebench\_reasoning\_prob\_prompt\_template]
\small
\textbf{Task:} Solve the following reasoning problem. Provide your best guess along with a confidence probability score (0.0 to 1.0).
\\\\
\textbf{Instructions:}\\
- Carefully read and analyze the problem.\\
- Reason through the solution step by step, if helpful.\\
- You might see several questions in the problem. You need to answer all of them and provide your final answer separated by commas.\\
- At the end, present your final answer and a confidence probability score in the following XML format:\\
<answer>final answer here</answer>\\
<confidence>confidence probability score here</confidence>\\
\\
\textbf{Example output:} \\
\char91 YOUR\_REASONING\char93\\
<answer>no, yes, no</answer>\\
<confidence>0.8</confidence>\\
\\
\textbf{Now, here is the problem:} \\
\{problem\}
\end{tcolorbox}

\begin{tcolorbox}[promptbox, title=self\_reflection\_aime\_prompt\_template]
\small
\textbf{Task:} Solve the following math problem.
\\\\
\textbf{Instructions:}\\
- Please reason step by step.\\
- At the end, present your final answer in the following XML format:\\
<answer>final answer here</answer>\\
\\
\textbf{Example output:} \\
\char91 YOUR\_REASONING\char93\\
<answer>123</answer>\\
\\
\textbf{Now, here is the problem:} \\
\{problem\}
\end{tcolorbox}

\begin{tcolorbox}[promptbox, title=self\_reflection\_mc\_prompt\_template]
\small
\textbf{Task:} Solve the following multiple-choice problem.
\\\\
\textbf{Instructions:}\\
- Carefully read and analyze the problem.\\
- Reason through the solution step by step, if helpful.\\
- At the end, present your final answer in the following XML format:\\
<answer>your final answer here</answer>\\
\\
\textbf{Example output:} \\
\char91 YOUR\_REASONING\char93\\
<answer>A</answer>\\
\\
\textbf{Now, here is the problem:} \\
\{problem\}
\end{tcolorbox}

\begin{tcolorbox}[promptbox, title=self\_reflection\_simpleqa\_prompt\_template]
\small
\textbf{Task:} Solve the following QA problem.
\\\\
\textbf{Instructions:}\\
- Carefully read and analyze the problem.\\
- Reason through the solution step by step, if helpful.\\
- At the end, present your final answer in the following XML format:\\
<answer>final answer here</answer>\\
\\
\textbf{Example output:} \\
\char91 YOUR\_REASONING\char93\\
<answer>123</answer>\\
\\
\textbf{Now, here is the problem:} \\
\{problem\}
\end{tcolorbox}

\begin{tcolorbox}[promptbox, title=self\_reflection\_livebench\_reasoning\_prompt\_template]
\small
\textbf{Task:} Solve the following reasoning problem.
\\\\
\textbf{Instructions:}\\
- Carefully read and analyze the problem.\\
- Reason through the solution step by step, if helpful.\\
- At the end, present your final answer in the following XML format:\\
<answer>final answer here</answer>\\
\\
\textbf{Example output:} \\
\char91 YOUR\_REASONING\char93\\
<answer>no, yes, no</answer>\\
\\
\textbf{Now, here is the problem:} \\
\{problem\}
\end{tcolorbox}

\begin{tcolorbox}[promptbox, title=reflection\_prompt\_template]
\small
\textbf{Task:} Reflect on the following problem and solution, and provide a final confidence score to the solution.
\\\\
\textbf{Instructions:}\\
- Carefully read and analyze the problem and solution.\\
- Reason through the solution step by step, if helpful.\\
- At the end, present your final answer in the following XML format:\\
<confidence>confidence score here</confidence>\\
\\
\textbf{Example output:} \\
\char91 YOUR\_REASONING\char93\\
<confidence>80\%</confidence>\\
\\
\textbf{Now, here is the problem and solution:} \\
\textbf{Problem:}\\
\{problem\}\\\\
\textbf{Solution:}\\
\{solution\}
\end{tcolorbox}

\section{More Analyses of Factuality Benchmarks}
\label{appendix:further_factuality}
As noted in the main text, our evaluated SFT reasoning models are fine-tuned from base models without general-purpose RL. To further examine the impact of initialization, we also evaluated two additional SFT reasoning models, OpenThinker2-32B \cite{guha2025openthoughtsdatarecipesreasoning} and R1-Distill-Llama-70B \cite{deepseekai2025deepseekr1incentivizingreasoningcapability}, both of which are fine-tuned from instruction-tuned checkpoints rather than base models. Their results are presented in \autoref{tab:not_attempted_appendix}. These findings indicate that the original trend persists: SFT reasoning models fine-tuned from instruction-tuned checkpoints continue to exhibit significantly lower “\textit{not attempted}” rates as those initialized from base models.

\begin{table}[htbp]
\centering
\resizebox{0.8\columnwidth}{!}{%
\begin{tabular}{lcc}
\toprule
Model Size & Instruct & SFT Reasoning \\
\midrule
32B & 2492 & 43 \\
70B & 1107 & 78 \\
\bottomrule
\end{tabular}
}
\caption{The total number of “\textit{not attempted}” responses in SimpleQA and FreshQA. These SFT reasoning models are trained from instruction-tuned checkpoints.}
\label{tab:not_attempted_appendix}
\end{table}

\section{Hyperparameters}
Our inferential hyperparameters are shown in \autoref{app:hyper}.
\begin{table*}[h!]
\centering
\begin{tabular}{lccc}
\toprule
\textbf{Model} & \textbf{Temperature} & \textbf{top\_p} & \textbf{Seq.Len} \\
\midrule
Qwen2.5-14B-Instruct & 0.7 & 0.8 & 3K \\
Qwen2.5-32B-Instruct & 0.7 & 0.8 & 3K\\
DeepSeek-V3 & 0.6 & 1.0 & 8K \\
DeepSeek-R1-Distill-Qwen-14B & 0.6 & 0.95 & 16K \\
DeepSeek-R1-Distill-Qwen-32B & 0.6 & 0.95 & 16K \\
DeepCoder-14B-Preview & 0.6 & 0.95 & 16K\\
Skywork-OR1-32B-Preview & 0.6 & 0.95 &  16K \\
DeepSeek-R1 & 0.6 & 0.95 & 16K \\
\bottomrule
\end{tabular}
\caption{Summary of hyperparameters used for each evaluated model (temperature, top\_p, and generation sequence length).}
\label{app:hyper}
\end{table*}

\section{GenAI Statement}
The authors used Cursor for coding support and ChatGPT for writing revisions as needed.

\section{License Discussion}
In all evaluated models, DeepSeek‑released variants (e.g., R1, V3‑0324) are licensed under the MIT License, and Qwen‑related models follow the Qwen License (Tongyi Qianwen License Agreement or Apache‑2.0 where applicable). Both license types permit academic research use, and our use of these models complies with their licensing terms.

\end{document}